\documentclass[fleqn]{article-hermes}

\journal{\LOBJET. Volume 8 -- n°2/2005}{1}{15}

\title[Mode d'emploi de \textit{article-hermes.cls}]%
      {Recommandation mobile, sensible au contexte de contenus évolutifs: Contextuel-$\epsilon$-Greedy}

\author{Djallel Bouneffouf}

\address{%
Department of Computer Science, Télécom SudParis, UMR CNRS Samovar, 91011 Evry 
Cedex, France\\
Djallel.Bouneffouf@it-sudparis.eu}

\resume{ 
La plupart des systèmes de recommandation sensibles au contexte se concentrent davantage sur la recommandation de documents pertinents à un utilisateur tout en intégrant des données contextuelles sans pour autant considérer le problème de l'évolution des contenus. Nous présentons dans cet article un algorithme qui non seulement recommande des documents adaptés à la situation de l'utilisateur mais aussi qui intègre le fait que ces documents sont susceptibles d'évoluer au cours du temps. Nous modélisons la recommandation personnalisée de documents comme un problème de bandit contextuel, une approche dans laquelle un algorithme d'apprentissage choisit de nouveaux  articles en se basant sur des données contextuelles de l'utilisateur. Le défit est de maintenir automatiquement  l'équilibre entre l'exploitation et exploration et ce en fonction de la situation de l'utilisateur. 
Nous avons effectué des évaluations avec des données réelles issues d'un journal d' événements d'une société informatique. Les résultats expérimentaux montrent que notre algorithme améliore les résultats des algorithmes étudiés.
}

\abstract{Most existing approaches in Mobile Context-Aware Recommender Systems focus on recommending relevant items to users taking into account contextual information, such as time, location, or social aspects. However, none of them has considered the problem of content dynamicity.
We introduce in this paper an algorithm that tackles this dynamicity. It is based on dynamic exploration/exploitation tradeoff and can adaptively balance the two aspects by deciding which user's situation is most relevant for exploration or exploitation.  We conduct evaluations with real online event log data from a computer science compagny. The experimental results demonstrate that our algorithm outperforms surveyed algorithms.
}

\motscles{système de recommandation, exploration/exploitation, apprentissage automatique}


\begin{document}


\maketitlepage

\section{Introduction}
Les technologies mobiles ont accès à une vaste collection d'information, n'importe où et n'importe quand. 
En particulier, la plupart des utilisateurs mobiles professionnels acquièrent et  maintiennent une grande quantité de documents dans leur référentiel. En ce sens, les systèmes de recommandation doivent rapidement identifier l'importance des documents nouveaux, tout en s'adaptant à la perte de valeur des documents anciens. Dans un tel contexte, il est crucial d'identifier et de recommander un contenu intéressant pour les utilisateurs.
\\
Les systèmes de recommandation représentent un thème de recherche fondamental en plein essor. 
L'objectif de ces systèmes est de filtrer un flux entrant d'informations (documents) de façon personnalisée pour chaque utilisateur, tout en s'adaptant en permanence au besoin d'information de chacun. En d'autres termes, il s'agit de présenter à l'utilisateur des contenus en lien avec ses préférences et ses attentes mais qu'il n'aurait pas consulté spontanément. 
Actuellement, il existe deux grandes approches de filtrage : le filtrage basé sur le contenu, où le profil décrit les caractéristiques de contenu que les documents intéressants sont censés présenter, et le filtrage collaboratif, où
le profil est constitué de l'historique des évaluations émises par l'utilisateur sur des documents, sans qu'aucun élément de contenu ne soit rendu explicite. Dans le premier cas, le système compare les nouveaux documents au profil de chaque utilisateur et recommande ceux qui sont les plus proches, et dans le second, le système compare les utilisateurs entre eux sur la base de leurs évaluations passées pour créer les communautés, et chaque utilisateur reçoit les documents jugés intéressants par sa communauté. Chacune de ces approches présente des avantages et des inconvénients.  Dans le filtrage basé sur le contenu, les documents recommandés sont toujours semblables aux documents déjà sélectionnés  par l'utilisateur \cite{Ml}. Le filtrage collaboratif quant à lui, ne propose une recommandation intéressante que si le chevauchement entre les historiques est important et que l'ensemble des contenus est relativement statique \cite{Sc}. Pour pallier ses inconvénients, de nombreux systèmes, dits « hybrides », les combinent de sorte que l'incapacité du filtrage collaboratif de recommander de nouveaux documents est réduite en le combinant avec un filtrage basé sur le contenu \cite{Li}.

Dans un cadre mobile, d'autres aspects sont à considérer. D'une part, le contexte de l'utilisateur influence le choix de l'information pertinente; d'autre part, la taille des écrans des dispositifs mobiles rend les interactions avec l'utilisateur plus difficiles. Les contenus étant dynamiques et susceptibles de changer fréquemment (ajouts, suppressions), le filtrage basé sur le contenu et le filtrage collaboratif ne sont pas adaptés \cite{Ch}.
Une nouvelle famille de systèmes de recommandation appelés Systèmes de Recommandation Sensibles au Contexte (SRSC) ou « context-aware » se développe de plus en plus dans la littérature comme par exemple \cite{Ba}  \cite{ci}, \cite{co} 
qui se basent sur le contexte de l'utilisateur en modélisant des données comme sa localisation, l'heure et les personnes aux alentours. La principale limite de ces approches est qu'elles ne tiennent pas compte de la dynamicité des contenus. 

Une façon d'appréhender le problème des systèmes de recommandation, à savoir présenter à l'utilisateur des contenus pertinents, est d'utiliser l'apprentissage automatique pour faire des prédictions sur des données non précédemment vues en apprenant de l'expérience passée. Ce problème s'apparente au problème du bandit puisqu'il peut se traduire comme un compromis entre exploration et exploitation: dans la phase d'exploration, le but est d'obtenir des récompenses pour améliorer la qualité des prédictions; dans la phase d'exploitation, le but est de présenter des items avec la plus haute évaluation avec les informations obtenues  jusqu'ici. 
Un algorithme de bandit B exploite son expérience passée pour sélectionner les documents qui apparaissent plus fréquemment. En outre, ces documents apparemment optimaux, peuvent s'avérer sous-optimaux en raison de l'imprécision dans la connaissance de B.
Afin d'éviter ce cas indésirable, B peut explorer les documents en choisissant des documents apparemment sous-optimaux afin de recueillir de plus amples informations à leur sujet. On peut dire que l'exploitation diminue la satisfaction à court terme de l'utilisateur puisque certains documents sous-optimaux peuvent être choisis. Cependant, l'obtention d'informations sur les documents avec des récompenses moyennes peut affiner l'estimation de la pertinence de ces derniers et  augmenter ainsi la satisfaction de l'utilisateur à long terme.
\\
De toute évidence, ni une pure exploration ni une pure exploitation ne donne un résultat satisfaisant. Trouver un bon compromis entre les deux est donc un véritable défit. Une solution classique au problème de bandit est la stratégie $\epsilon$-greedy \cite{Ev}. Avec la probabilité 1-$ \epsilon $, cet algorithme choisit les meilleurs documents sur la base des connaissances actuelles, et, avec la probabilité $\epsilon$, il choisit de manière uniforme tous les autres documents. Le paramètre $\epsilon$ contrôle essentiellement le compromis exr/exp. L'inconvénient de cet algorithme est qu'il est difficile de déterminer à l'avance la valeur optimale de $\epsilon$. 

Pour adresser ce problème, nous proposons dans ce papier un algorithme appelé contextuel-$ \epsilon $-greedy \cite{Bouneffouf2012} qui permet d'atteindre cet objectif en équilibrant de manière adaptative le compromis exr/exp en fonction de la situation de l'utilisateur. Cet algorithme étend la stratégie $\epsilon $-greedy avec une mise à jour de l'exr/exp, en sélectionnant des situations de l'utilisateur appropriées que ce soit pour l'exploration ou pour l'exploitation.
La suite du papier est organisée comme suit. La Section \ref{etatdelart} examine des travaux connexes. La Section \ref{notions} donne les notions clés utilisées dans le présent papier. La Section 4 présente notre modèle SRSC et décrit les algorithmes impliqués dans la démarche proposée. L'évaluation expérimentale est illustrée dans la Section 5. La dernière section conclut le papier et donne des directions possibles pour les travaux futurs.

\section{\'{E}tat de l'art} \label{etatdelart}
Dans ce qui suit, nous présentons une vue d'ensemble des algorithmes de bandit, puis nous aborderons les systèmes de recommandation sensibles au contexte récents qui considèrent  la situation de l'utilisateur dans la recommandation.

\subsection{Vue d'ensemble des algorithmes de bandit ($ \epsilon $-greedy)}
Le problème d'exploration vs exploitation ou "bandit à bras multiples" a été initialement décrit par Robbins \cite{cir}. La stratégie $\epsilon$-greedy est la plus utilisée pour résoudre le problème de bandit et a d'abord été décrite dans \cite{Wa}. Le $\epsilon$-greedy choisit un document aléatoirement avec une fréquence $ \epsilon $, sinon il choisit le document avec la plus haute récompense moyenne estimée. Le paramètre $ \epsilon $ doit être dans l'intervalle fermé [0, 1] et son choix est laissé à l'utilisateur.
\\
La première variante de la stratégie $ \epsilon $-greedy est ce que \cite{Ev}, et \cite{Ma} appellent la stratégie $ \epsilon $-beginning. Cette stratégie permet l'exploration une seule fois au début de l'interaction. Pour un nombre $I$ donné d'itérations, les documents sont tirés au hasard au cours des premières $ \epsilon I$ itérations. Durant les autres $(1- \epsilon )I$ itérations, le document avec la plus haute moyenne estimée est tiré.
\\
Une autre variante de la stratégie $ \epsilon $-greedy est ce que 
\cite{Ma} appelle la stratégie $\epsilon$-decreasing. Dans cette stratégie, un document est tiré au hasard avec une fréquence $ \epsilon_i$, où $i$ est l'indice de l'itération actuelle. La valeur de l'exploration $ \epsilon_i$ est donnée par $ \epsilon_i$ = {$ (\epsilon_0  / i$)} où $ \epsilon_0 $ $ \in $[0,1]. Le document avec la plus haute moyenne estimée est tiré avec une fréquence de $(1-\epsilon_i)$. Quatre autres stratégies sont présentées dans \cite{Au}. Ces stratégies ne sont pas décrites ici car  les expériences faites par \cite{Au}  montrent que, en utilisant des paramètres choisis avec soin, $ \epsilon $-decreasing est toujours aussi bonne que les autres stratégies.
\\ 
Contrairement au problème de bandit standard, où un ensemble fixe d'actions est possible, dans les SRSC, des documents anciens peuvent expirer ou être modifiés et de nouveaux documents peuvent  apparaître. Par conséquent, il n'est pas souhaitable d'explorer d'un coup, au début, comme dans \cite{Ev}, ou de diminuer de façon monotone l'effort d'exploration, comme dans \cite{Ma}.
\\
Une solution pourrait être la prise ne compte du contexte de l'utilisateur dans l'équilibre exr/exp. Peu de travaux de recherche sont dédiés à l'étude du problème du bandit contextuel dans les systèmes de recommandation. Dans \cite{Li1}, les auteurs étendent la stratégie $\epsilon $-greedy en mettant à jour dynamiquement la valeur d'exploration $ \epsilon $. \`A chaque itération, ils lancent une procédure d'échantillonnage pour sélectionner un $ \epsilon $ à partir d'un ensemble fini de candidats. Les probabilités associées aux candidats sont initialisées uniformément et mises à jour avec le gradient exponentiel (EG) \cite{Ki}. Cette mise à jour augmente la probabilité d'un candidat $ \epsilon $ si elle conduit à un clic de l'utilisateur. Comparée à la fois à $ \epsilon $-beginning et à $ \epsilon $-decreasing, cette technique améliore les résultats.\\
Dans \cite{Li}, les auteurs modélisent la recommandation comme un problème de bandit contextuel. Ils proposent une approche dans laquelle un algorithme d'apprentissage sélectionne séquentiellement des documents pour servir les utilisateurs sur la base d'informations extraites des logs des utilisateurs. Pour maximiser le nombre total de clics des utilisateurs, ce travail propose l'algorithme de calcul LINUCB qui est efficace.
\\
Les auteurs de \cite{Au}, \cite{Ev}, \cite{Ma}, \cite{Li}, \cite{Li1} décrivent une façon intelligente d'équilibrer l'exploration et l'exploitation. Cependant, aucun d'entre eux ne considère la situation de l'utilisateur pour la recommandation.

\subsection{Gestion de la situation de l'utilisateur}
Peu de travaux de recherche sont dédiés à la gestion de la situation de l'utilisateur dans les systèmes de recommandation mobiles. Dans \cite{Be}, \cite{Sa}, les auteurs proposent une méthode qui consiste à construire un profil utilisateur dynamique en se basant sur le temps et l'expérience de l'utilisateur. Les préférences de l'utilisateur sont pondérées dans son profil en fonction de la situation (temps, localisation) et son comportement. Pour modéliser l'évolution des préférences de l'utilisateur 
au cours du temps, (e.g., journée de travail, vacances), les concepts dans le profil de l'utilisateur sont pondérés pour chaque nouvelle expérience. L'activité et le profil de l'utilisateur sont utilisés ensemble pour recommander des contenus pertinents.
\\
 \cite{Ra} décrit un fonctionnement de SRSC sur trois dimensions du contexte qui se complètent mutuellement pour être très ciblées. Tout d'abord, le SRSC analyse des informations telles que les carnets d'adresses des clients pour évaluer le niveau d'affinité sociale entre les utilisateurs. Puis, il combine l'affinité sociale avec les dimensions spatio-temporelles et  l'historique de l'utilisateur dans le but d'améliorer la qualité des recommandations.
\\
Dans \cite{Sp}, les auteurs présentent une technique pour effectuer un filtrage collaboratif. Le dispositif mobile de chaque utilisateur stocke toutes les évaluations explicites faites par son propriétaire ainsi que les notes provenant d'autres utilisateurs. Seuls les utilisateurs se trouvant à proximité spatio-temporelle sont en mesure d'échanger des notes et ils montrent comment ceci offre un filtrage naturel à base de contextes sociaux.
\\
Chaque article cité ci-dessus  recommande aux utilisateurs des informations intéressantes en fonction de la situation de l'utilisateur, mais ils ne considèrent pas l'évolution des contenus. 

Nous traitons dans cet article précisément le problème de la dynamicité des contenus en prenant en compte la situation de l'utilisateur dans la stratégie exr/exp. De plus, notre intuition est que le niveau de criticité de la situation dans la gestion du compromis exr/exp améliore le résultat des SRSC. Notre approche  propose une exploration élevée lorsque la situation courante de l'utilisateur n'est pas critique et réalise une forte exploitation dans le cas inverse.
\\
Nous précisons que nous avons choisi d'améliorer l'algorithme $\epsilon $-greedy de part son fort compromis exr/exp; cependant, notre approche peut être utilisée pour adapter n'importe quel algorithme de bandit dans un environnement contextuel. 

\section{Notions clés}\label{notions}
Dans cette section, nous présentons brièvement les notions clés qui sont utiles dans nos explications tout au long du document.
\begin{description}
\item[Le contexte de l'utilisateur] Le contexte est modélisé par un triplet $C$ = ($O_{Localisation}$, $ O_{Temps}$, $O_{Social}$)  où chaque $O_{i}$, $i\in$ \{$Localisation, Temps, Social$\}, est une ontologie qui gère le type de données d'une dimension du contexte. Nous nous concentrons sur ces trois dimensions car elles couvrent toutes les informations nécessaires à notre cas d'étude. 
\subsubsection*{Localisation}
Il y a différentes manières de caractériser la localisation retournée par le GPS. En effet, la localisation est une position géographique à base de coordonnées qui peut être définie par une adresse ou par une information plus sémantique permettant de caractériser le type de lieu, e.g., bureau, restaurant, etc.
Nous utilisons une API commerciale (Google maps) qui permet de passer d'une localisation à base de coordonnées à une adresse; puis, en utilisant l'ontologie Localisation que nous avons développée, 
nous passons de l'adresse au type de lieu. Dans la Fig. \ref{fig:location} nous pouvons voir un court extrait de l'ontologie développée sous protégé, nous permettant, par exemple, de savoir que \'Evry et Paris sont localisées en Ile de France, ou bien de passer d'une adresse particulière à un nom d'entreprise.  

\begin{center}
\begin{figure}[!h]
\begin{center}
  \includegraphics[width=0.75\textwidth]{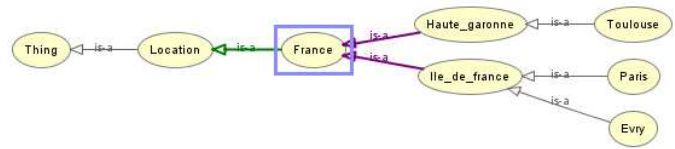}\\
\caption{Ontologie Localisation}
\label{fig:location}       
\end{center}
\end{figure}
\end{center}

\subsubsection*{Temps}
Les informations temporelles sont complexes, continues et peuvent être représentées avec différents niveaux de granularité. Pour définir l'aspect temporel caractérisant la situation de l'utilisateur, nous avons abstrait le temps par des périodes utiles dans notre domaine d'application (e.g., jour de travail, vacances, pause déjeuner).  \\
Par exemple, pour la valeur temporelle "Mon Oct 3 12:10:00 2011", l'ontologie Temps permet d'inférer qu'il s'agit d'un "jour de travail". La Fig. \ref{fig:temps} montre un extrait de l'ontologie Temps. 

\begin{center}
\begin{figure}[!h]
\begin{center}
  \includegraphics[width=0.5\textwidth]{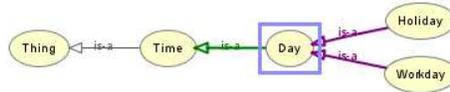}\\
\caption{Ontologie Temps}
\label{fig:temps}       
\end{center}
\end{figure}
\end{center}

\subsubsection*{Relations sociales}
Les relations sociales d'un utilisateur se réfèrent à la typologie des interlocuteurs potentiels de l'utilisateur, comme par exemple ses collègues, un client ou son manager.\\ 
L'ontologie sociale que nous avons construite, dont un extrait est montré dans la Fig. \ref{fig:sociale}, permet de décrire le réseau social professionnel de l'utilisateur à travers des concepts et des propriétés. 

\begin{center}
\begin{figure}[!h]
\begin{center}
  \includegraphics[width=0.75\textwidth]{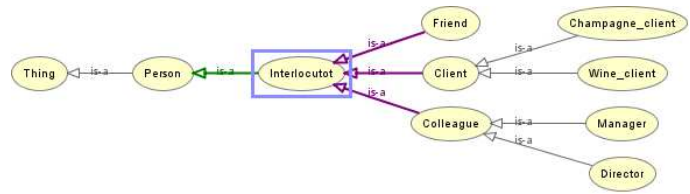}\\
\caption{Ontologie sociale}
\label{fig:sociale}       
\end{center}
\end{figure}
\end{center}

\item [La situation de l'utilisateur] Une situation est une instanciation du contexte de l'utilisateur. Elle est modélisée sous forme d'un triplet S = ($O_{Localisation}.x_i $, $O_{Temps}.x_j $, $O_{Social}.x_k$) où $x_i$, $x_j$ et $x_k$ sont des concepts ontologiques ou des instances en fonction de la granularité de l'information. Supposons que les données suivantes sont détectées à partir du téléphone mobile de l'utilisateur : l'information spatiale donnée par le GPS indique la latitude et la longitude d'un point (48.89, 2.23), la date est 01\_Oct\_2012 et l'agenda indique "rencontre avec Paul Gerard". La situation correspondante est : S = ((48.89, 2.23), 01\_Oct\_2012, Paul\_Gerard). \`A partir de ces données de bas niveau, nous utilisons les ontologies Localisation, Temps et Sociale pour donner plus de sémantique à la situation, par exemple, en déduire que l'utilisateur "est\_en\_réunion" avec un client ou "est\_au\_restaurant" avec un Financier. Nous interprétons ensuite le comportement de l'utilisateur dans les situations ainsi décrites.

Parmi l'ensemble des situations visées, certaines d'entre elles sont caractérisées comme étant des situations critiques pour la recommandation.

\item [Les Situations Critiques:]
 
L'ensemble de situations critiques (SC) constitue une classe de situations où l'utilisateur a besoin de la plus optimale recommandation. C'est le cas, par exemple, lors d'une réunion professionnelle. Dans une telle situation, le système doit effectuer exclusivement de l'exploitation, l'exploration étant axée sur l'apprentissage.
Dans les situations considérées comme non critiques, (par exemple, "est\_à\_la\_maison" ou "est\_en\_vacances" avec des amis), le système peut  se permettre de faire de l'exploration en recommandant des informations qui ne sont pas forcément très liées aux intérêts de l'utilisateur rencontrées dans des situations précédentes. Les SC sont prédéfinies par l'expert du domaine. Dans notre cas d'étude mené avec des commerciaux (étude décrite dans la Section 5), les commerciaux considèrent comme étant des SC, par exemple, un déjeuner d'affaire avec un client,  S1 = (restaurant, midi, client), ou être en réunion avec son manager,  S2 = (société, matin, manager). 

\item [Les préférences utilisateur:]Les préférences utilisateur sont déduites au cours de ses navigations, par exemple le nombre de clics sur les documents visités ou le temps passé sur un document. Soit \textit{UP} l'ensemble des préférences soumises dans le système par un utilisateur  pendant une situation donnée. Chaque document dans \textit{UP} est représenté comme un vecteur unique \textit{d} = ($ c_{1}$, ..., $c_{n}$), où $c_{i}$ ($i$ = 1, .., n) est la valeur d'un élément caractérisant les préférences par rapport à \textit{d}. Nous considérons les éléments suivants : le nombre total de clics sur \textit{d}, le temps total consacré à la lecture de \textit{d} et le nombre de fois que \textit{d} a été recommandé.

\item [Le modèle utilisateur:]Le modèle utilisateur est structuré comme une base de cas composée d'un ensemble de situations et les préférences utilisateur associées. Ce modèle est notée U = {($S^{i}$; $UP^{i}$)}, où $S^{i}$ est la situation de l'utilisateur et $UP_{i}$ ses préférences.

\end{description}

\section{Le Modèle SRSC Proposé}
Comme nous l'avons mentionné dans la Section \ref{Introduction}, la recommandation de documents est modélisée comme un problème de bandit contextuel qui contient des informations sur la situation de l'utilisateur. Formellement, un algorithme de bandit procède à un ensemble d'essais $e_t$, $t = 1 ... T$. Pour chaque essai $e_t$, l'algorithme effectue les tâches suivantes:

Tâche 1 : Soit $S^t$ la situation courante de l'utilisateur et $PS$ l'ensemble des situations passées. Le système compare  $S^t$ aux situations $S^c$ dans $PS$ afin de choisir celle qui est la plus proche, $S^{p}$, comme indiqué dans l'Eq. \ref{sitPlusSim}. 

\begin{equation}\label{sitPlusSim}
S^p=argmax_{S^{c}\in PS}(sim(S^t,S^c))
\end{equation}                                                   

Inspirée de \cite{Bou11}, la métrique de similarité sémantique est calculée avec l'Eq. \ref{similariteSit}.  

\begin{equation}\label{similariteSit}
sim(S^t, S^c)=\sum_{j}\alpha_{j}sim_{j}(x^t_j,x^c_j))
\end{equation}         

Dans l'Eq. \ref{similariteSit}, $sim_j$ est la mesure de similarité liée à la dimension $j$ entre deux concepts $x_{j}^{t}$ et $x_j^c$; $\alpha_{j}$ est le poids associé à la dimension $j$ (dans la phase expérimentale de notre travail, $\alpha_{j}$ a une valeur de 1 pour toutes les dimensions, i. e. nous considérons que toutes les dimensions du contexte ont le même poids). Cette similarité dépend de la distance entre $x_j^t$ et $x_j^c$ dans l'ontologie correspondante. Nous utilisons la même mesure de similarité que \cite{Ml} définie par l'Eq. \ref{similariteConcept}.

\begin{equation}\label{similariteConcept}
sim_{j}(x_{j}^{t},x_{j}^{c})=2*\frac{deph(LCS(x_j^t, x_j^c))}{deph(x_j^t)+ deph(x_j^c)}
\end{equation}  \\                     

Dans l'Eq. \ref{similariteConcept}, LCS (Least Common Subsumer) est l'ancêtre commun le plus proche entre $x_j^t$ et $x_j^c$;  {\em deph} est le nombre de n{\oe}uds sur le chemin entre le noeud en cause et la racine de l'ontologie.

Tâche 2: Soit $D$ la collection de documents et $D^p \in D$, l'ensemble des documents recommandés dans $S^p$. Après avoir récupéré $S^p$, le système observe le comportement de l'utilisateur lors de la lecture de chaque document $d \in D^p$. En se basant sur les récompenses observées, l'algorithme choisit le document $d_p$ avec la récompense $r_p$ la plus élevée et le recommande à l'utilisateur.

Tâche 3: Après avoir reçu la nouvelle récompense de l'utilisateur, l'algorithme améliore sa stratégie de sélection de documents avec la nouvelle observation: à la situation $S^t$, le document $d_p$ obtient une récompense $r_t$.

Lorsqu'un document est présenté à l'utilisateur, si celui-ci le sélectionne par un clic, une récompense de 1 est donnée au système pour ce document; sinon, la récompense est 0. La récompense pour un document est précisément son taux de clics, Click Through Rate (CTR). Le CTR est le ratio entre le nombre de clics pour un document et le nombre de fois qu'il a été recommandé.

\subsection{L'algorithme $ \epsilon $-greedy}

L'algorithme $ \epsilon $-greedy recommande un nombre $N$ prédéfini de documents sélectionnés en utilisant l'Eq. \ref{eq:egreedy}.  
 
\begin{equation}        \label{eq:egreedy}                                                                        
  d_i = \left\{ \begin{array}{rcl}
             argmax_{d \in D^p - \{d_1, ..., d_i-1\}}(getCTR(d))& if & q > \epsilon \\ 
             Random(D^p - \{d_1, ..., d_i-1\})     & otherwise
             \end{array}\right.     
\end{equation}                                       

Dans l'Eq. \ref{eq:egreedy}, $i \in {1, ... , N}$; $D^p = {d_1, ..., d_p}$ ($p > N$) est l'ensemble des documents correspondant aux préférences de l'utilisateur; $getCTR()$ calcule le CTR d'un document; $Random()$ retourne un élément aléatoire d'un ensemble donné, ce qui permet d'effectuer l'exploration; $q$ est une valeur aléatoire, uniformément distribuée sur [0, 1], qui définit le compromis exr/exp; $ \epsilon $ est la probabilité de recommander un document par exploration aléatoire.

\subsection{L'algorithme contextuel-$ \epsilon $-greedy}
\label{contextuelegreedy}

Pour une meilleure adaptation de l'algorithme $ \epsilon $-greedy à des situations critiques, l'algorithme contextuel-$ \epsilon $-greedy compare la situation courante $S^t$ de l'utilisateur avec la classe SC de situations critiques. En fonction de la similarité entre la situation courante $S^t$ et la situation la plus similaire $S^m\in$ SC, $B$ étant le seuil de similarité (discuté dans la Sec. \ref{evaluation}), deux scénarios sont possibles :

(1) Si $sim (S^t, S^m) \ge B$, la situation courante est critique; l'algorithme $ \epsilon $-greedy est utilisé avec $ \epsilon $ = 0 (exploitation) et $S^t$ est insérée dans la classe SC de situations critiques.

(2) Si $sim (S^t, S^m) < B$, la situation courante n'est pas critique; l'algorithme $ \epsilon $-greedy est utilisée avec $ \epsilon > 0$ (exploration) calculé comme indiqué dans Eq. \ref{eq:epsilon}.

\begin{equation}   \label{eq:epsilon}                                                                             
 \epsilon =\left\{
 \begin{array}{rcl}1-(\frac{sim(S^t,S^m)}{B}) & if & sim(S^t,S^m) < B\\
   0    &    otherwise
 \end{array}\right.      
\end{equation}        

Pour résumer, le système ne fait pas d'exploration lorsque la situation courante de l'utilisateur est critique; dans le cas contraire, le système explore. Le degré d'exploration diminue lorsque la similarité entre $S^t$ et $S^m$ augmente. 

\section{\'{E}valuation}\label{evaluation}

Afin d'évaluer empiriquement les performances de notre approche, et en l'absence d'un cadre d'évaluation standard, nous proposons un cadre d'évaluation fondé sur un ensemble d'entrées de journal d'événements. Les principaux objectifs de l'évaluation expérimentale sont les suivants: 

\begin{itemize}
\item trouver la valeur du seuil optimal $B$ décrite dans la section \ref{contextuelegreedy};

\item évaluer les performances de l'algorithme contextuel-$ \epsilon $-greedy proposé.
\end{itemize} 

Dans ce qui suit, nous décrivons  l'ensemble de nos données expérimentales, puis nous présentons et discutons les résultats obtenus.

Nous avons mené une étude avec la collaboration de Nomalys\footnote{Nomalys est une société qui fournit une application graphique permettant aux utilisateurs de smartphones d'accéder aux données de leur entreprise.} société française de logiciels. Cette société fournit une application qui stocke pour chaque utilisateur un historique des situations comprenant l'heure, la localisation et les interlocuteurs (information sociale). L'étude a pris 18 mois et a généré 178 369 situations. \\
Pour chaque entrée, les données saisies sont remplacées par des informations plus abstraites à l'aide des ontologies temporelle, spatiale et sociale (cf. Section \ref{notions}). De l'étude du journal, nous avons obtenu un total de 2 759 283 entrées relatives à la navigation des utilisateurs, exprimées avec une moyenne de 15.47 entrées par situation.

\subsection*{Calcul du seuil de similarité} 

Afin de définir la valeur du seuil de similarité, nous utilisons une classification manuelle comme point de départ et nous comparons les résultats avec ceux obtenus par notre technique. Nous prenons un échantillon aléatoire d'environ 0,1\%  des  situations et nous groupons les situations similaires manuellement; ensuite nous comparons les groupes construits avec les résultats obtenus par notre algorithme de similarité, en utilisant des valeurs de seuil différentes.

\begin{figure}[h]
\begin{center}
 \includegraphics[width=0.75\textwidth]{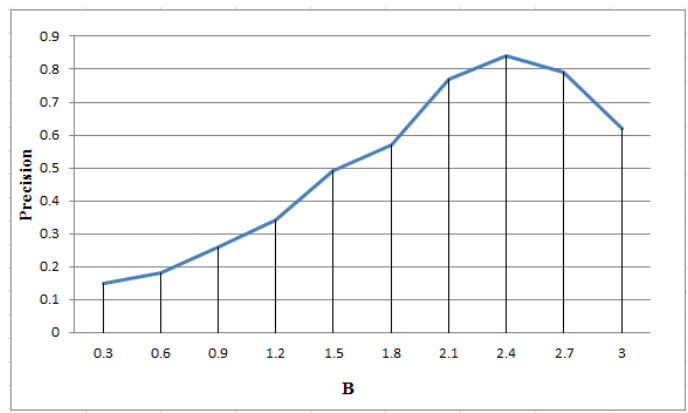}
 \caption{Impact du seuil $B$ sur la précision.} \label{fig:threshold}
\end{center}

\end{figure}

La Fig. \ref{fig:threshold} montre l'effet de la variation du seuil de similarité $B$ dans l'intervalle [0, 3] sur la précision globale. Les résultats montrent que la meilleure performance est obtenue lorsque $B$ a la valeur 2,4, atteignant une précision de 0,849. Par conséquent, nous utilisons la valeur du seuil optimal $B = 2,4$ pour tester nos algorithmes.

\subsection*{Comparaison des algorithmes}

\begin{figure}[h]
\begin{center}
 \includegraphics[width=0.95\textwidth]{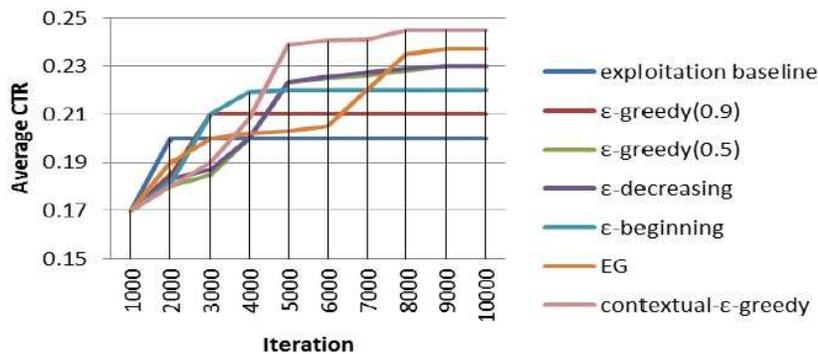}
 \caption{Le CTR moyen des algorithmes d'exr/exp.} \label{fig:evaluationiteration}
\end{center}
\end{figure}

Pour comparer l'algorithme contextuel-$ \epsilon $-greedy avec $ \epsilon $-greedy, nous avons échantillonné 10000 situations aléatoirement.
\\
L'étape de test consiste à évaluer les algorithmes de test pour chaque situation en utilisant le taux de clics moyen. Le CTR moyen pour une itération donnée est le rapport entre le nombre total de clics et le nombre total de recommandations faites par le système. Nous avons calculé le CTR moyen toutes les 1000 itérations.\\ 
Le nombre $N$ de documents retourné par le système de recommandation pour chaque situation est de 10 et nous avons exécuté la simulation jusqu'à ce que le nombre d'itérations atteigne 10000, qui est le nombre d'itérations où tous nos algorithmes convergent.\\
Dans la première expérience, en plus d'une pure exploitation de base, nous avons comparé notre algorithme avec les algorithmes décrits dans les travaux connexes (Sec. \ref{etatdelart}) : $ \epsilon $-greedy; $ \epsilon $-beginning, $ \epsilon $-decreasing et EG. Dans la Fig. \ref{fig:evaluationiteration}, l'axe horizontal représente le nombre d'itérations et l'axe vertical représente la métrique de performance.

Nous avons paramétré les différents algorithmes comme suit : $\epsilon$-greedy a été testé avec deux valeurs de $\epsilon$: 0,5 et 0,9; $ \epsilon $-decreasing et EG utilisent le même ensemble {$ \epsilon_i$ = 1 - 0,01 * i, i = 1, ..., 100 }; $\epsilon$-decreasing utilise au début la valeur la plus élevée et elle décroit de 0,01 toutes les 100 itérations, jusqu'à atteindre la plus petite valeur.
\\
 L'ensemble des algorithmes testés a de meilleures performances que la ligne de base (exploitation pure). Toutefois, pour les 2000 premières itérations, l'exploitation pure a une convergence plus rapide. Mais dans le long terme, tous les algorithmes exr/exp améliorent le CTR moyen jusqu'à la convergence.
 \\
Nous avons plusieurs observations concernant les différents algorithmes d'exr/exp. Pour l'algorithme $ \epsilon $-decreasing, la convergence en CTR moyen augmente à mesure que diminue le $ \epsilon $ (l'exploration augmente).
\\
Pour le $\epsilon$-greedy(0,9) et le $ \epsilon$-greedy(0,5), même après convergence, les algorithmes donnent encore respectivement 90\% et 50\% d'occasions aux documents ayant le CTR moyen faible d'être sélectionnés, ce qui diminue de manière significative leurs résultats.
\\
Alors que l'algorithme contextuel-$\epsilon $-greedy converge avec un taux de CTR moyen élevé, sa performance globale n'est pas aussi bonne que $ \epsilon $-decreasing. Son taux de CTR moyen est faible dans les premières itérations en raison de plus d'exploration, mais il converge plus rapidement. L'algorithme contextuel-$\epsilon $-greedy apprend effectivement le meilleur $\epsilon $, il a le meilleur taux de convergence, augmente le CTR moyen par un facteur de 2 par rapport à la ligne de base et surpasse tous les algorithmes étudiés.
L'amélioration provient d'un compromis dynamique entre exr/exp, contrôlé par la prise en compte des SC. Au stade précoce, cet algorithme tire pleinement parti de l'exploration sans perdre des occasions d'établir de bons résultats.

\section{Conclusion}
Dans cet article, nous avons étudié le problème de l'exploration et de l'exploitation dans des systèmes de recommandation mobiles sensibles au contexte et nous avons proposé une nouvelle approche qui concilie, de façon adaptative, exr/exp en considérant la situation de l'utilisateur.
\\
Afin d'évaluer les performances de l'algorithme proposé, nous l'avons comparé avec d'autres algorithmes standards d'exr/exp. Les résultats expérimentaux montrent que notre algorithme est plus performant sur le taux de CTR moyen dans différentes configurations.
\\
Pour la suite de ce travail, nous prévoyons d'évaluer notre algorithme embarqué dans un dipositif mobile et de le tester sur d'autres jeux de données publiques afin d'augmenter la reproductibilité de l'étude.

\nocite{*}
\bibliography{exemple-biblio}



\end{document}